\documentclass[a4paper]{article}
\usepackage{graphicx}
\usepackage{multirow}

\usepackage{INTERSPEECH2022}
\usepackage[utf8]{inputenc}
\usepackage{tikz,booktabs,float,color,adjustbox}

\title{Adaptive multilingual speech recognition with pretrained models}

\name{Ngoc-Quan Pham$^1$~~Alex Waibel$^1$$^,$$^2$~~Jan Niehues$^1$}
\address{
  $^1$Interactive Systems Lab, Karlsruhe Institute of Technology, Karlsruhe, Germany\\
      $^2$Carnegie Mellon University, Pittsburgh PA, USA}
\email{ngoc.pham@kit.edu}

\begin{document}
\maketitle
\begin{abstract}
Multilingual speech recognition with supervised learning has achieved great results as reflected in recent research. With the development of pretraining methods on audio and text data, it is imperative to transfer the knowledge from unsupervised multilingual models to facilitate recognition, especially in many languages with limited data. Our work investigated the effectiveness of using two pretrained models for two modalities: wav2vec 2.0 for audio and MBART50 for text, together with the adaptive weight techniques to massively improve the recognition quality on the public datasets containing CommonVoice and Europarl. Overall, we noticed an 44\% improvement over purely supervised learning, and more importantly, each technique provides a different reinforcement in different languages. We also explore other possibilities to potentially obtain the best model by slightly adding either depth or relative attention to the architecture. 
\end{abstract}


\section{Introduction}

The sequence-to-sequence approach is widely used in speech recognition (SR) nowadays~\cite{park2019specaugment,pham2019transformer,gulati2020conformer}, and many research works are dedicated to show that their capabilities relying on a single architecture often match or are even better than traditional hybrid or CTC systems with separately optimized components~\cite{nguyen2019improving,gulati2020conformer}. Moreover, this approach is also successfully realized in multilingual speech recognition, allowing one model to contain the information of many languages shared in one single body of neural networks~\cite{pratap2020massively}.

Despite such promising potential, this approach inevitably requires abundant training data because it combines acoustic models, language models and alignment into the same model~\cite{bahdanau2016end,chan2016listen}. In fact, the labeled data necessary for the task is limited, and even more limited outside of English and other mainstream languages, making building a reliable speech recognizer for many languages even more challenging. One can even question if sequence-to-sequence is a sustainable approach without fully utilizing data as well as the hybrid systems. 

On the other hand, the availability of unlabeled data is virtually unlimited, and more importantly they can also be available under the form of pretrained models. Since the release of Bidirectional Transformers or BERT~\cite{devlin2018bert} with masked language model pretraining and its success in applying for natural language processing, many Transformers~\cite{vaswani2017attention} based architectures have been the silver bullet to tackle low-resource problems. Recently, in the speech domain, Transformers are also well adopted in supervised learning~\cite{pham2019transformer,pham2020relative} and unsupervised pretraining with contrastive predicting methods~\cite{schneider2019wav2vec,baevski2020wav2vec}. Most importantly, these models can be effortlessly trained on a multilingual dataset and acquire the acoustic or syntactic information of many languages. Replacing the components in an e2e model with the pretrained ones is trivial and can potentially gain largely for multilingual recognition.

With the pretrained models available in our arsenal, in this paper we set out to explore the possibilities of combining them for multilingual ASR. While the application wav2vec 2.0 and MBART50 individually is ubiqutuous and the combination of them has been observed in Speech Translation~\cite{li2020multilingual}, to the best of our knowledge this is the first attempt in multilingual ASR, especially in a large scale with many languages with different data size.

More importantly, there are many improvements for the Transformers over the years that improve either modeling long range dependencies in self-attention~\cite{dai-etal-2019-transformer,pham2020relative} or adaptive components for multilingual models~\cite{bapna2019simple,pham2021efficient}. It is promising to also combine these techniques with the pretrained models for the best results.  

With such motivation, we carried out experiments with 32 languages ranging from high to very low resource and explore the possibilities of using pretrained models. Our contribution is as follows:
\begin{itemize}
    \item First, using multilingual pretrained acoustic and language models for the encoder and decoder respective brings a large improvement as a whole. However each pretrained module has a different influence on different languages, namely encoder pretraining is more impactful for languages with higher resources while the decoder counterpart is more effective for languages with medium-low resources. Surprisingly, many languages with extremely low resource (less than 5 hours) do not benefit much from this combination.
    \item Second, the language specific modulation techniques such as language adapters~\cite{bapna2019simple} and factorized adaptive weights~\cite{pham2021efficient} complement the two pretrained modules very well and have a strong impact on especially the low-resource languages mentioned above.
\end{itemize}

Moreover, there are different possibilities to integrate knowledge from pretrained models, and it is not necessary by simply replacing components. We also provided further analysis to the architecture, by showing that there are benefits to either improving the self-attention mechanism by adding relative positions during fine-tuning, or stacking the MBART encoders to the wav2vec counterpart.  

This is the continuation of the line of work building multilingual ASR systems based on sequence-to-sequence neural networks~\cite{kannan2019large,pratap2020massively,pham2021efficient}. Our work is available for public at\textit{ https://github.com/quanpn90/NMTGMinor} providing a highly CUDA-optimized implementation for both wav2vec and MBART which is potentially useful for the community.

\section{Modeling}
Transformers~\cite{vaswani2017attention} are a class of sequence-to-sequence models with an encoder and a decoder with attention. Both components are equipped with self-attention layers being able to well handle long-short range dependencies in the input, output or alignment between sequences. These properties make this class of model appeal for Speech Recognition~\cite{pham2019transformer,gulati2020conformer}.


\subsection{Pretrained Acoustic representation}
Transformer encoders can be used to learn useful representation from input masking and construction which is demonstrated in masked language model~\cite{devlin2018bert}. Following this trend, they are inevitably applied for audio signals, starting from learning to reconstruct the log-mel frequency features~\cite{jiang2019improving} to using quantization to learn latent variables~\cite{schneider2019wav2vec}.

In our work, the wav2vec 2.0 model~\cite{baevski2020wav2vec} is selected to replace the typically randomly initialized encoder. It consists of three main components: a convolutional feature extractor that convolves and downsamples the raw audio input, a deep Transformer encoder that learn high level representation from the downsample sequence, and a quantization module to generate latent variables. During pre-training, the network optimizes for a contrastive loss function while masking the speech input. wav2vec 2.0 showed that it can outperform other semi-supervised learning approaches using finetuning with a CTC model.

\subsection{Pretrained Multilingual BART}
BART~\cite{lewis2019bart} was designed to learn syntactic features in languages using Transformers. The network is tasked to reconstruct a particular sentence at the decoder given a noisy version at the encoder side. MBART~\cite{liu2020multilingual} and its extension MBART50~\cite{tang2020multilingual} took the BART training scheme and applied to 50 languages. This proved to be very useful for multilingual  translation by finetuning the network on parallel data. 

In speech recognition, data scarcity is even more problematic when the number of sentences fall short compared to other natural language tasks. The presence of MBART in the decoder is promising, especially for the self-attention parts which play the role of a language model. 
    
\subsection{Language Adaptive Components}
The development of multilingual models for either machine translation, speech translation or speech recognition often concern between versatility versus specialization~\cite{le2021lightweight}. The motivation comes from the disbelief that there are features being shared between languages and at the same time each language requires to selectively represented, and networks are encouraged to change "modes" depending on the language being processed~\cite{hampshire1992meta}. Since then, multingual model designers opt to use specific network components being presented for each language, ranging from weight generator~\cite{platanios-etal-2018-contextual} to  adapters~\cite{bapna2019simple,le2021lightweight} or recently adaptive weights adding scales and biases to each weight matrix in the whole architecture~\cite{pham2021efficient}. In this paper, the last two options are selected for investigation thanks to being computationally manageable.

On the one hand, Adapters plugged into pretrained models were introduced in computer vision~\cite{rebuffi2017learning} and later natural language processing~\cite{houlsby2019parameter} and recently in Transformers for text/speech translation~\cite{bapna2019simple,le2021lightweight}. They are materialized with a small multilayer perceptrons (MLP) with one hidden layer that acts as a \textit{downsampler} (for parameter efficiency). This MLP is serialized at the end of each layer in the Transformer to help the network changes the feature distribution based on languages. 
    

On the other hand, adaptive weights~\cite{pham2021efficient} was proposed based on the observation that neural networks evolve rapidly yet the core remains to be matrix multiplication. Therefore, it is possible to separate the weight matrix into a shared component $W_S$ and language dependent \textit{adaptive} scale $W_{ML}$ and bias $W_{BL}$. The simple matrix multiplication $Y = WX$ becomes:

\begin{equation}
    Y = (W_S \cdot W_{ML} + W_{BL})^\top X 
\end{equation}

In order to encourage the model to share parameters as well as keeping the parameters efficient, the adaptive weights are \textit{factorized} by using the form of 1-rank matrices~\cite{wen2020batchensemble} which can be compactly represented as a dot-product between two vectors. 
This factorization can be established with $k$ vectors per language so that there are $k$ independent weight factors followed by a summation, which increases the rank of the additional weight matrices.

\begin{equation}
    \label{eq:factorize}
    \bar{W} = \sum_i^k r_i s_i^\top
\end{equation}
Equation~\ref{eq:factorize} applies for all scale and bias matrices in the network.

Comparing two approaches, the advantage of the adapters is that they can increase the depth of representation in the network thanks for having an activation function in the downsampled layer. In contrast, the adaptive weights have the advantage to directly affect each layer function, such as the QKV-projection layer in self-attention, instead of applying a new function on the output of the layer. In the Transformers specifically this particular combination between a pretrained encoder and a decoder, the cross-attention layers in the decoder are left with weights untrained to connect two modalities audio and text. By being able to directly alter this function, the advantage of the adaptive weights are even more considerable. 

\subsection{Related Work}
Using pretrained models is very effective for low-resource machine translation~\cite{liu2020multilingual,tang2020multilingual} and recently speech translation~\cite{li2020multilingual} which can even handle zero translation from speech. In ASR, the presence of pretrained acoustic models~\cite{baevski2020wav2vec,hsu2021hubert} allows recognition to be possible with very little data. The combination of pretrained acoustic and language models are recently investigated via learning to relax the modality mismatch~\cite{zheng2021wav,wang2020bridging} yet still requires CTC and limited in a monolingual setup. In terms of multilingual ASR, there are various results in training a single model~\cite{pratap2020massively} that can overcome the monolingual performance. In this paper, we combined most prominent techniques to improve the results for many languages.

\section{Experiments}
There are 32 languages We report the error rates on the test sets of CommonVoice and Europarl. It is notable that both of the pretrained model (wav2vec) and the available supervised training data are mostly read speech, leading to the curiosity about the performance on a more spontaneous setting. 

Our speech recognition experiments are conducted using the public dataset including CommonVoice~\cite{ardila2019common} and Europarl~\cite{iranzo2020europarl} as training data. 

For the progressive comparison, we trained a competitive supervised model using the Transformer large configuration~\cite{vaswani2017attention} with 24 encoder layers, 8 decoder layers and relative attention~\cite{pham2020relative}. For transfer learning, we used the wav2vec 2.0 model pretrained with $53$ languages~\cite{babu2021xls} with the large configuration that has the same hidden size with our initial model. It is notable that the data used in pretraining is heavily biased to read speech including CommonVoice and Multilingual Librispeech~\cite{pratap2020mls}. For pretrained language models, MBART50~\cite{tang2020multilingual} with the same hidden size is used. For language specific modules, we use adapters with hidden layer size $512$ and adaptive weights with $k=8$ for bias and $k=1$ for scale matrices, so that they have the same number of additional parameters per language. For training, we used an effective batch size of around 2.84 hours of data per update, together with a linear decay learning rate that peaks at $0.001$\footnote{The decaying equation follows the same in Attention is all you need~\cite{vaswani2017attention}}. The supervised model takes $150K$ updates to converge, while the model with transfer learning takes $50K$ updates.\footnote{Training was possible using 4 NVIDIA A100 GPUs.} 

The performance impact of the pretrained modules are fully presented in Table~\ref{tab:result32}. The test data here is the combination of CommonVoice and Europarl wherein the latter is available. Since the languages widely vary in terms of data size, we divide them into three groups that have less than 10 hours (very low), between 10-100 hours (low) and more than 100 hours (medium-large) of training data. Notably, our supervised model outperformed the previously reported error rates~\cite{pham2021efficient, hou2021exploiting}. 

\begin{table*}[htb]
\caption{Performance(WER$\downarrow$) on the CommonVoice-Europarl dataset. Models include Transformers (\textbf{TF}), with wav2vec pre-training (\textbf{W}), with wav2vec and MBART50 (\textbf{WM}), with adapters (\textbf{WMA}) and factorized weights (\textbf{WMF}) and the version w/ frozen pt. weights.}
\label{tab:result32}

	\centering
	\begin{tabular}{lcccccccc}
	
		\toprule
        \textbf{Language} & Hours & \textbf{TF} & \textbf{W} & \textbf{WM} & \textbf{WMA} &\textbf{WMF}&\textbf{FWMA}&\textbf{FWMF}\\
        \midrule
         (de) & 1050 & 10.4    & 8.1  & 7.8  & 7.7 & \textbf{7.2} & 11.8 & 10.0     \\
         (nl) & 150  & 13.2    & 8.4  & 7.7  & 7.4 & \textbf{6.8} & 10.2 & 9.0     \\
         (fr)&  800  & 15.2    & 12.7 & 12.12 & 11.62 & \textbf{11.2} & 16.7 & 15.1  \\
         (it) & 325  & 11.5    & 8.7  & 8     & 7.8  & \textbf{6.5}  & 12.8 & 10.5  \\
         (fa) & 293  & 5.5	    & 4.8	& 3.9	& 4.2 &  \textbf{4.0} & 6.4 & 7.2       \\
         (pl) & 145  & 11.5    & 10.5  & 9.2	& 9.1 & \textbf{7.6} & 14.8    & 12.2 \\
         (pt) & 120  & 14.0    & 10.9 & 10.0	& 9.5 & \textbf{6.0} & 18.5   & 12.8 \\
         (es) & 400  & 10.9	& 8.0	& 7.6	& 7.4  & \textbf{6.2} &  11.2   & 9.8   \\
         (ru) & 148  & 10.0  	& 8.7   & 5.5	    & 5.7  & \textbf{5.4}  & 20.1   &  10.1\\
         (ta) & 198  & 28.6	& 24	& 20.2	& \textbf{20.7} & 21.0 & 31.4 & 31.5 \\
         (th) & 133  & 2.8	& 2.6	& 3.3	& 3.4 & \textbf{3.2} & 5.1 & 4.5 \\
         \midrule
         Average &  & 12.1 & 10.3 & 9.5 & 9.3 & \textbf{8.6} & 15.8 & 13 \\
         \midrule
         (ro) & 45   & 18.1    & 21.2 & 15.8 & 13.7 & \textbf{10.12} & 42.0 & 15.7 \\
         (ar) & 85   & 21.1	& \textbf{15.8}  & 18.7	& 17.6 & 18.2 & 31.2   & 23.2 \\
         (et) & 32   & 30.4	& 22.1	 & 14.8	& 15.1 & \textbf{13.2} & 32.5   & 21.5  \\
         (ja) & 26   & 13.0	& 10.3 	& 8.3	& 7.91 & \textbf{7.9} & 21.8 & 11.5 \\
         (zh) & 63   & 25.9	& 16.7 	& 15.2	& \textbf{14.6} & 14.7 & 37.6 & 18.2 \\
         (cs) & 49   & 19.8	& 15.8 	& 10.0	& 9.2 & \textbf{8.4} & 16.4 & 12.7 \\
         (lt) & 16   & 43.3	& 37.9	 & 31.9	& 26.7 & \textbf{25.5} & 71.3   & 30.0 \\
         (tr) & 30   & 10.4	& 13.6  & 7.5	    & 8.4  & \textbf{7.5} & 9.8 & 10.1     \\
         (id) & 23   & 14.0 & 13.6	& 7.5	  & 7.5 & \textbf{6.6} & 9 & 8.7 \\
         (mn) & 12   & 49.8	& 35.1	& 26.2	  & 26.0 & \textbf{24.3} & 90.4 & 32.0 \\
         (sv) & 35   & 24.7	 & 20.6	& 14.3	& 13.0  & \textbf{12.3} & 17.5 & 16.3 \\
         (uk) & 56   & 14 & 13.4	& 7.6	& 8.3 & \textbf{7.4} & 11.4 & 14.8 \\
         \midrule
         Average &  & 23.7 & 20 & 14.5 & 13.7 & \textbf{12.5} & 32.7 & 17.4 \\
         \midrule
         (lv) & 6    & 41.9    & 57.0	 & 41.01	& \textbf{22.3}  & \textbf{22.3} & 79.1   & 25.0 \\
         (vi) & 3    & 49.5 &  53.5	& 46.1	& 35.6	  & \textbf{34.0} & 104.1 & 38.3    \\
         (ka) & 6    & 58.6	& 48.0	& 48.3	  & 33.0 & \textbf{32.09} & 130.3 & 39.0 \\
         (sl) & 9    & 20.5	& 26.5 	& 14.6	& 10.4 & \textbf{9.1} & 10.5 & 12.8 \\
         (fi) & 6    & 54.2	& 48.5  & 41.0	& 31.4 & \textbf{30.0} & 109.2  & 35.2 \\
         (hi) & 8    & 46.1    & 46.4	 & 36.4	& 28.6 & \textbf{27.6} & 99.5   & 31.1 \\
         (gl) & 7    & 26.5	& 15.3 	& 15.3	& 11.2 & \textbf{9.8} & 48.8 & 16.0 \\
         (ur) & 0.6  & 78.0	& 68.2 	& 62.3	& \textbf{56.6} & 57.0 & 104.1 & 70.0 \\
         (kk) & 0.73 & 86.5    & 76.8	 & 86.4     & \textbf{60.6}	& 61.0 & 104.3  & 70.03 \\
        \midrule
        Average &  & 51.3 & 48.9 & 43.5 &  32.2 & \textbf{31.4} & 87.8 & 37.5 \\
        \midrule
        Overall &  & 30 & 24.8 & 20.9 & 17.4 &  \textbf{16.6} & 41.6 & 21.6 \\
		\bottomrule
	\end{tabular}
\end{table*}

\subsection{Impact of acoustic pretraining}
Compared to the Transformer model without any pretraining (\textbf{TF}), having the encoder pretrained with XLS-R (\textbf{W}) brings a substantial improvement to the average error rates by $18\%$, and this enables many languages to reach $10\%$ errors or lower. Across the three groups however, there is a clear difference in impact. While the high-medium group enjoys a $20\%$ decrease in error rate, this figure drops to $16\%$ in the first group, and the third group is only improved by $4\%$. 
\\
While this is rather surprising, compared to the previously reported of wav2vec 2.0 pretrained models on very low resource settings~\cite{baevski2020wav2vec}, it is explainable by the difference of the approaches used in their and our works. The pretrained acoustic model has often been used directly with the CTC loss function to generate characters which heavily requires an external language model. In our setting, we are limited in both acoustic and text resources for the languages in the third group, and data scarcity makes learning to align from attention~\cite{chan2016listen} even harder. 

\subsection{Effects from pretrained language model}
With that observation, initializing the decoder with the MBART pretrained model is expected to alleviate the data scarcity problem. In fact, comparing the model with two pretrained modules (\textbf{WM}) with the previous one showed a benefit of $15\%$ error reduction for the former. 

Some languages have worse performance with the MBART decoder, such as Arabic, Turkish or Thai. This can be explained as a negative effect of the large byte-pair encoding~\cite{sennrich2016bpeacl} model shared between many languages originally used with MBART training. A large vocabulary size of $250K$ allow for large granularity which is far from characters or phonemes, the units that speech recognition models are often trained upon. Nevertheless, this is apparently not a problem for most languages.

Analyzing the individual performance of each group, the medium-large group is not impressively improved with just a $5\%$ reduction. This result shows that having an additional six layers of decoders and even with pretrained weights only has a minimal effect on the result and network depth has a clearer impact on the source side than the target side in end-to-end speech recognition~\cite{pham2019transformer}. Probably the model does not struggle with the test data in terms of syntax, despite the fact that the text data here is not comparable to typical language models. Nevertheless, the second group receives a clearer merit from the pretrained language model, by a $27\%$ improvement, and totally $39\%$ improvement compared to the original multilingual Transformer. Even with a mismatched pretrained weights for the cross-attention module, it is still a noticeable improvement coming from the pretrained self-attention, feed-forward and layer normalization weights. The languages benefiting the most are Romanian, Estonian, Czech, Turkish, Indonesian, Swedish and Ukrainian.  

The effect on the third group is modest at $11\%$ reduction and the error rates remain very high for Urdu (ur), Kazakh (kk), Finnish (fi), Latvian (lv) or Vietnamese (vi). Many of the languages are also syntactically with many morphological word forms, such as Finnish, making recognition even more challenging. The most surprising improvement, however, comes from Slovenian that massively reduces from $26\%$ to $14.6\%$. The overall struggling mainly comes from data scarcity which is not adequate for the decoder cross-attention layers. 

\subsection{Effect of the language-specific modules} 

As can be seen from Table~\ref{tab:result32}, both techniques are able to help the model generalize better in all language groups. Most importantly, the very low resource group witnesses $27\%$ and $26\%$ improvement using adaptive weights and adapters respectively, compared to the baseline model with two pretrained modules.


In order to quantify their impact, we proceeded to freeze all of the pretrained parameters and only fine-tuned the language-specific parameters. There is a difference in how each technique handles this situation. With the presence of only adapters, the errors in all languages deteriorate rapidly in all language groups. Many languages in the low-group experience very bad results including Romanian, Arabic, Chinese, Lithuanian and especially many members within the very-low group exceed $90\%$ error rates. While this is unexpected, we can see that the main problem here is the cross-attention layer which is not familiar with the inputs coming from two modalities. The adapters are not able to drive the bad context vectors (weighted-sum of the encoder inputs) into meaningful representation in the low resource condition.

The adaptive factorized weights do not have this problem because they directly alter cross-attention. As a result, the performance is much better than the adapters even though they still fall behind the baseline without language-specific modules.  

\subsection{Further analysis}
In the previous sections, we presented the most important enhancement for our multilingual setup coming from the pretrained modules and the language adaptive components. The success of using the language adaptive components in various places, either breaking the \textit{layer dynamics} with adapters or the \textit{function dynamics} with the adaptive weights suggests that further improvement can be found by adding more information to the system instead of treating the pretrained model as an immovable black box.

Here we follow the implementation in~\cite{pham2020relative} to add relative positions into the wav2vec model, by adding the content-position interaction together with the content bias and position bias to self-attention to all self-attention layers of wav2vec with factorized weights. The additive information allowed for the reduction of the average error rate to $16.04\%$ (3\% improvement) and especially an $12.7\%$ on the large-medium group. This evidently helps the model learns better with adequate data.

Surprisingly, an alternative attempt to enhance the encoder simply by stacking the MBART50 encoder on top of the wav2vec encoder (without any length conversion) yields similar results, compared to the baseline \textbf{WM}, it improves the large-medium group by an impressive $18.3\%$, yet only $3.2\%$ overall. Not only does stacking the encoder increase the encoder depth, it also helps the cross-attention layers because they are familiar with the output of the text encoder during training. Training a stacked model with the adaptive techniques would probably result in the best model in our experiments, however it was unfortunately beyond our computational tolerance. 

\section{Conclusion}
In this work, we massively improved multilingual ASR using a combination of three techniques: encoder pretraining, decoder pretraining and adaptive weights. Our empirical results indicate that each technique has a different impact on different languages varying in linguistic characteristic and data size. Nevertheless, at the very low condition the model still struggles to reach an acceptable error rate. Our analysis in using adaptive weights shed light on the future work, in which multimodal pretraining~\cite{ao2021speecht5} is potentially beneficial to address the cross-attention layers. 

\section{Acknowledgements}

This work was realized within the project ELITR which has received funding from the European Unions Horizon 2020 Research and Innovation Programme under grant agreement No 825460.

\bibliographystyle{IEEEtran}

\bibliography{mybib}
\newpage

\end{document}